\pdfoutput=1

\documentclass[11pt]{article}

\usepackage{acl}
\usepackage{times}
\usepackage{latexsym}

\usepackage[T1]{fontenc}

\usepackage[utf8]{inputenc}
\usepackage{algorithm}
\usepackage{algorithmic}

\usepackage{amsmath}
\usepackage{graphicx}
\usepackage{booktabs}
\usepackage{bbm}
\usepackage{subcaption}
\usepackage{tabularx,ragged2e}
\usepackage{multicol,multirow}
\usepackage{enumitem}
\usepackage{soul}
\usepackage{microtype}

\usepackage{xcolor}

\title{Unsupervised Natural Language Inference Using PHL Triplet Generation}

\author{Neeraj Varshney,~~ 
  Pratyay Banerjee,~~ 
  Tejas Gokhale,~~ 
  Chitta Baral
  \\
  Arizona State University \\
  \texttt{\{nvarshn2, pbanerj6, tgokhale, cbaral\}}@asu.edu
  }

\begin{document}
\maketitle
\begin{abstract}
Transformer-based models achieve impressive performance on numerous Natural Language Inference (NLI) benchmarks when trained on respective training datasets.
However, in certain cases, training samples may not be available or collecting them could be time-consuming and resource-intensive.
In this work, we address the above challenge and present an explorative study on unsupervised NLI, a paradigm in which no human-annotated training samples are available.
We investigate it under three settings: \textit{PH, {P}}, and \textit{{NPH}} that differ in the extent of unlabeled data available for learning.
As a solution, we propose a procedural data generation approach that leverages a set of sentence transformations to collect PHL (Premise, Hypothesis, Label) triplets for training NLI models, bypassing the need for human-annotated training data.
Comprehensive experiments with several NLI datasets show that the proposed approach results in accuracies of up to $ 66.75\%, 65.9\%,  65.39\%$ in PH, P, and NPH settings respectively, outperforming all existing unsupervised baselines.
Furthermore, fine-tuning our model with as little as ${\sim}0.1\%$ of the human-annotated training dataset ($500$ instances) leads to $12.2\%$ higher accuracy than the model trained from scratch on the same $500$ instances.
Supported by this superior performance, we conclude with a recommendation for collecting high-quality task-specific data.
\end{abstract}

\section{Introduction}
\begin{figure}
    \centering 
    \small
    \includegraphics[width=0.85\linewidth]{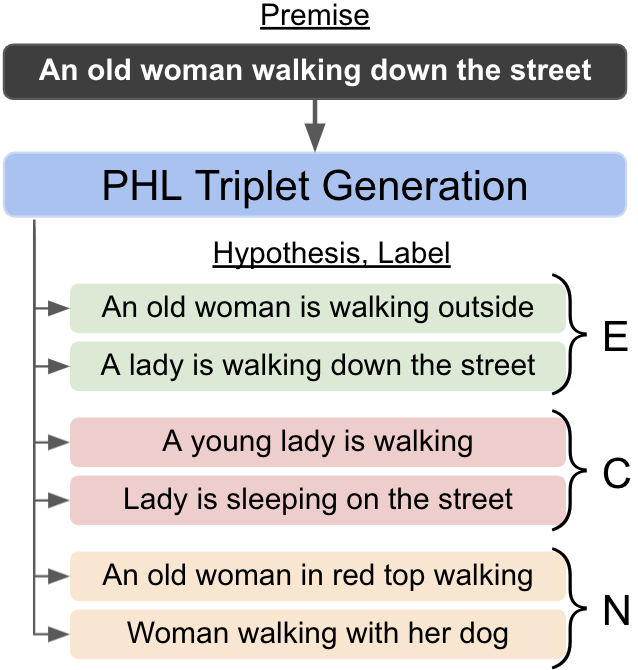}
    \caption{Illustrating our procedural data generation approach for unsupervised NLI. 
    A sentence is treated as premise, and multiple hypotheses conditioned on each label (Entailment- E, Contradiction- C, and Neutral- N) are generated using a set of sentence transformations. 
    }
    \label{fig:teaser_figure}
\end{figure}
\begin{figure*}
    \centering 
    \small
    \includegraphics[width=0.92\linewidth]{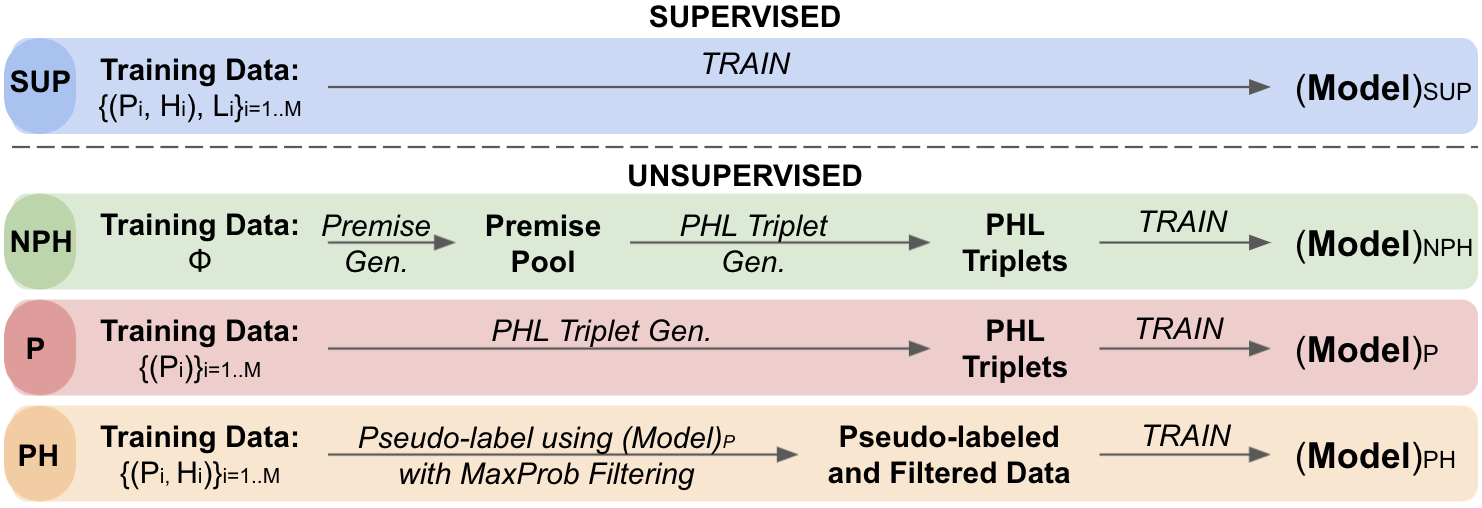}
    \caption{
    Comparing supervised NLI with our three unsupervised settings.
    For unsupervised settings, we procedurally generate PHL triplets to train the NLI model.
    In \textbf{NPH setting}, a premise pool is collected from raw text corpora such as Wikipedia and then used for generating PHL triplets.
    In \textbf{P setting}, we directly apply these transformations on the available premises.
    In \textbf{PH setting}, we leverage the P-setting model to pseudo-label and filter the provided unlabeled PH pairs and then train the NLI model using this pseudo-labeled dataset.
    }
    \label{fig:all_settings_figure}
\end{figure*}
Natural Language Inference (NLI) is the task of determining whether a ``hypothesis'' is true (Entailment), false (Contradiction), or undetermined (Neutral) given a ``premise''.
State-of-the-art models have matched human performance on several NLI benchmarks, such as SNLI \cite{bowman-etal-2015-large}, Multi-NLI \cite{williams-etal-2018-broad}, and Dialogue NLI \cite{welleck-etal-2019-dialogue}.
This high performance can be partially attributed to the availability of large training datasets; SNLI (570k), Multi-NLI (392k), and Dialogue-NLI (310k). 
For new domains, collecting such training data is time-consuming and can require significant resources.
What if no training data was available at all?

In this work, we address the above question and explore \textit{Unsupervised NLI}, a paradigm in which no human-annotated training data is provided for learning the task.
We study three different unsupervised settings: \textit{{PH}, {P}}, and \textit{{NPH}} that differ in the extent of unlabeled data available for learning.
In {PH}-setting, unlabeled premise-hypothesis pairs are available i.e. data without ground-truth labels.
In {P}-setting, only a set of premises are available i.e. unlabeled partial inputs.
The third setting {NPH} does not provide access to any training dataset, and thus it is the hardest among the three unsupervised settings considered in this work.

We propose to solve these unsupervised settings using a procedural data generation approach.
Given a sentence, our approach treats it as a premise (P) and generates multiple hypotheses (H) corresponding to each label (L = Entailment, Contradiction, and Neutral) using a set of sentence transformations (refer to Figure \ref{fig:teaser_figure}).
This results in creation of Premise-Hypothesis-Label (PHL) triplets that can be used for training the NLI model.
In the P and PH settings, we directly apply our sentence transformations over the available premises to generate PHL triplets.
However, in the NPH setting, premises are not available.
We tackle this challenge by incorporating a premise generation step that extracts sentences from various raw text corpora such as Wikipedia and short stories. We use these extracted sentences as premises to generate PHL triplets.
In Figure \ref{fig:all_settings_figure}, we compare the four settings (one supervised and three unsupervised) and show our approach to develop an NLI model for each setting.

To evaluate the efficacy of the proposed approach, we conduct comprehensive experiments with several NLI datasets.
We show that our approach results in accuracies of $ 66.75\%, 65.9\%$, and $65.39\%$ on SNLI dataset in PH, P, and NPH settings respectively, outperforming all existing unsupervised methods by ${\sim}$13\%.
We also conduct experiments in low-data regimes where a few human-annotated labeled instances are provided and show that further fine-tuning our models with these instances consistently achieves higher performance than the models fine-tuned from scratch.
For example, with just 500 labeled instances, our models achieve 8.4\% and 10.4\% higher accuracy on SNLI and MNLI datasets respectively.
Finally, we show that fine-tuning with `adversarial' instances instead of randomly selected human-annotated instances further improves the performance of our models; it leads to 12.2\% and 10.41\% higher accuracy on SNLI and MNLI respectively. \\

\noindent In summary, our contributions are as follows:
\begin{enumerate}[noitemsep]
    \item We explore three unsupervised settings for NLI and propose a procedural data generation approach that outperforms the existing approaches by ${\sim}13$\% and raises the state-of-the-art unsupervised performance on SNLI to $66.75$\%.
    
    \item We also conduct experiments in low-data regimes and demonstrate that further fine-tuning our models with the provided instances achieves $8.4\%$ and $10.4\%$ higher accuracy on SNLI and MNLI datasets respectively.
    
    \item Finally, we show that using `adversarial' instances for fine-tuning instead of randomly selected instances further improves the accuracy. It leads to $12.2\%$ and $10.41\%$ higher accuracy on SNLI and MNLI respectively. Supported by this superior performance, we conclude with a recommendation for collecting high-quality task-specific data.

\end{enumerate}
We release the implementation\footnote{\href{https://github.com/nrjvarshney/unsupervised_NLI}{https://github.com/nrjvarshney/unsupervised\_NLI}} of our procedural data generation approach and hope that our work will encourage research in developing techniques that reduce reliance on expensive human-annotated data for training task-specific models.

\section{Related Work}

\paragraph{Unsupervised Question-Answering:}
The \textit{unsupervised} paradigm where no human-annotated training data is provided for learning has mostly been explored for the Question Answering (QA) task in NLP.
The prominent approach involves synthesizing QA pairs and training a model on the synthetically generated data.
\citet{lewis-etal-2019-unsupervised, dhingra-etal-2018-simple, fabbri-etal-2020-template} propose a template-based approach, while \citet{puri-etal-2020-training} leverage generative models such as GPT-2~\cite{radford2019language} to synthesize QA pairs.
\citet{banerjee-baral-2020-self} create synthetic graphs for commonsense knowledge and propose knowledge triplet learning.
\citet{wang2021towards} leverage few-shot inference capability of GPT-3 \cite{NEURIPS2020_1457c0d6} to synthesize training data for SuperGLUE \cite{Wang2019SuperGLUEAS} tasks.
For visual question answering, \citet{gokhale2020vqa} use template-based data augmentation methods for negation, conjunction, and  \citet{banerjee-etal-2021-weaqa} utilize image captions to generate training data.
\citet{gokhale2021semantically} use linguistic transformations in a distributed robust optimization setting for vision-and-language inference models.

\paragraph{Unsupervised NLI:}
In NLI, \citet{cui-etal-2020-unsupervised} propose a multimodal aligned contrastive decoupled learning method (MACD) and train a BERT-based text encoder.
They assign a label (E, C, N) based on the cosine similarity between representations of premise and hypothesis learned by their text encoder.
Our approach differs from MACD as we leverage a procedural data generation step based on a set of sentence transformations and do not leverage data from other modalities.
We use MACD as one of the baselines in our experiments.



\section{Unsupervised NLI} 

In NLI, a premise-hypothesis pair $(P, H)$ is provided as input and the system needs to determine the relationship $L{\in}\{\textit{Entailment}, \textit{Contradiction}, \textit{Neutral}\}$ between $P$ and $H$.
In the \textbf{supervised setting}, a labeled dataset $D_{train}{=}\{(P_i, H_i),  L_i\}_{i=1}^M$ consisting of $M$ instances which are usually human-annotated is available for training.
However in the unsupervised setting, labels $L_i$ are not available, thus posing a significant challenge for training NLI systems.
Along with this standard unsupervised setting (referred to as PH), we consider two novel unsupervised settings (P and NPH) that differ in the extent of unlabeled data available for learning:
\paragraph{PH-setting:} It corresponds to the standard unsupervised setting where an unlabeled dataset of PH pairs ($\{(P_i, H_i)\}_{i=1}^M$) is provided.

\paragraph{P-setting:} In this setting, only premises from $D_{train}$ i.e ($\{(P_i)\}_{i=1}^M$) are provided. 
It is an interesting setting as the large-scale NLI datasets such as SNLI \cite{bowman-etal-2015-large} and MultiNLI \cite{williams-etal-2018-broad} have been collected by presenting only the premises to crowd-workers and asking them to write a hypothesis corresponding to each label.
Furthermore, this setting presents a harder challenge for training NLI systems than the PH-setting as only partial inputs are provided.

\paragraph{NPH-setting:} Here, no datasets (even with partial inputs) are provided. Thus, it corresponds to the hardest unsupervised NLI setting considered in this work.
This setting is of interest in scenarios where we need to make inferences on a test dataset but its corresponding training dataset is not available in any form.

From the above formulation, it can be inferred that the hardness of the task increases with each successive setting (PH$\rightarrow$P$\rightarrow$NPH) as lesser and lesser information is made available.
In order to address the challenges of each setting, we propose a two-step approach that includes a pipeline for procedurally generating PHL triplets from the limited information provided in each setting (Section \ref{pht_triptle_gen}), followed by training an NLI model using this procedurally generated data (Section \ref{training_nli_model}).
Figure \ref{fig:all_settings_figure} highlights the differences between four NLI settings (one supervised and three unsupervised) and summarizes our approach to develop an NLI model for each setting.

\section{PHL Triplet Generation}
\label{pht_triptle_gen}
To compensate for the absence of labeled training data, we leverage a set of sentence transformations and procedurally generate PHL triplets that can be used for training the NLI model.
In P and PH settings, we apply these transformations on the provided premise sentences.
In the NPH setting where premises are not provided, we extract sentences from various raw text corpora and apply these transformations on them to generate PHL triplets.

\subsection{$\mathcal{P}$: Premise Generation}
\label{premise_gen}
We extract sentences from raw text sources, namely, COCO captions~\cite{lin2014microsoft}, ROC stories~\cite{mostafazadeh-etal-2016-corpus}, and Wikipedia to compile a set of premises for the NPH setting.
We use these text sources as they are easily available and contain a large number of diverse sentences from multiple domains. 

\textbf{ROC Stories} is a collection of short stories consisting of five sentences each. We include all these sentences in our premise pool.
\textbf{MS-COCO} is a dataset consisting of images with five captions each. We add all captions to our premise pool.
From \textbf{Wikipedia}, we segment the paragraphs into individual sentences and add them to our premise pool.

We do not perform any sentence filtration during the premise collection process. However, each transformation (described in subsection \ref{transformations}) has its pre-conditions such as presence of verbs/adjectives/nouns that automatically filter out sentences from the premise pool that can not be used for PHL triplet generation.

\subsection{$\mathcal{T}$: Transformations} 
\label{transformations}
Now, we present our sentence transformations for each NLI label.
\begin{table*}[t]
    \small
    \centering
    \resizebox{\linewidth}{!}{
    \begin{tabular}{@{}p{0.13\linewidth}>{\RaggedRight}p{0.39\linewidth}>{\RaggedRight}p{0.39\linewidth}>{\RaggedRight}p{0.04\linewidth}}
    \toprule
        \textbf{Transformation} &
        \textbf{Original Sentence (Premise)} &
        \textbf{Hypothesis} &
        \textbf{Label}
        \\
    \midrule
              PA	& 
          Fruit and cheese sitting on a black plate	& 
          There is fruit and cheese on a black plate &
          E\\

        
        PA + ES + HS & 
        A large elephant is very close to the camera &
        Elephant is close to the photographic equipment &
        E\\
        
        CW-noun & 
        Two horses that are pulling a carriage in the street & 
        Two dogs that are pulling a carriage in the street &
        C\\ 
        
        CV & 
        A young man sitting in front of a TV & 
        A man in green jersey jumping on baseball field &
        C\\ 
        
        
        PA + CW &
        A woman holding a baby while a man takes a picture of them &
        A kid is taking a picture of a male and a baby &
        C\\

        
        FCon & 
        A food plate on a glass table & 
        A food plate made of plastic on a glass table &
        N\\ 
        
        PA + AM &
        Two dogs running through the snow &
        The big dogs are outside &
        N\\
    \bottomrule

    \end{tabular}
    }
    \caption{Illustrative examples of PHL triplets generated from our proposed transformations. E,C, and N correspond to the NLI labels Entailment, Contradiction, and Neutral respectively.}
    \label{tab:transforms_combined}
\end{table*}

Table \ref{tab:transforms_combined} illustrates examples of PHL triplets generated from these transformations.

\subsubsection{Entailment:}
In NLI, the label is entailment when the hypothesis must be true if the premise is true.
\paragraph{Paraphrasing (PA):} 
Paraphrasing corresponds to expressing the meaning of a text (restatement) using other words and hence results in entailment premise-hypothesis pairs.
We use the Pegasus \cite{zhang2019pegasus} tool to generate up to 10 paraphrases of a sentence and use them as hypothesis with the original sentence as the premise
\footnote{Further details are in Appendix Section \ref{supp_transformations} \label{footnote1}}.

\paragraph{Extracting Snippets (ES):}  We use dependency parse tree to extract meaningful snippets from a sentence and use them as hypothesis with the original sentence as the premise. 
Specifically, we extract sub-trees that form a complete phrase or a sentence. 
For example, from the sentence ``\textit{A person with red shirt is running near the garden}'', we create entailing hypotheses ``\textit{A person is running near the garden}'', ``\textit{A person is running}'', ``\textit{A person is near the garden}'', etc.
We implement $10$ such techniques using spacy \cite{spacy}\textsuperscript{\ref{footnote1}}.

\paragraph{Hypernym Substitution (HS):} A hypernym of a word is its supertype, for example, ``animal'' is a hypernym of ``dog''.
We use WordNet~\cite{miller1995wordnet} to collect hypernyms and replace noun(s) in a sentence with their corresponding hypernyms to create entailment hypothesis. 
For example, from the premise ``\textit{A black dog is sleeping}'', we create ``\textit{A black animal is sleeping}''. 
Note that swapping the premise and hypothesis in this case gives us another PH pair that has a `Neutral' relationship.

\paragraph{Pronoun Substitution (PS):} Here, we leverage Part-of-Speech (POS) tagging of spacy to heuristically substitute a noun with its mapped pronoun. 
For example, substituting ``boy'' with ``he'' in the sentence ``\textit{boy is dancing in arena}'' results in an entailing hypothesis ``\textit{he is dancing in arena}''\textsuperscript{\ref{footnote1}}.

\paragraph{Counting (CT):} Here, we count nouns with common hypernyms and use several templates such as ``\textit{There are \{count\} \{hypernym\}s present''} to generate entailing hypotheses.
For instance, from the sentence ``\textit{A motorbike and a car are parked}'', we create hypothesis ``\textit{Two automobiles are parked}''.
We also create contradiction hypotheses using the same templates by simply changing the $count$ value such as ``\textit{There are five automobiles present}''\textsuperscript{\ref{footnote1}}.

\subsubsection{Contradiction:}
The label is contradiction when the hypothesis can never be true if the premise is true. 

\paragraph{Contradictory Words (CW):} We replace noun(s) and/or adjective(s) (identified using spacy POS tagging) with their corresponding contradictory words. For example, replacing the word `big' with `small' in ``\textit{He lives in a big house}'' results in a contradictory hypothesis ``\textit{He lives in a small house}''.
For contradictory adjectives, we collect antonyms from wordnet and for nouns, we use the function `$most\_similar$' from gensim \cite{rehurek2011gensim}
\textsuperscript{\ref{footnote1}}.

\paragraph{Contradictory Verb (CV):} 
We collect contradictory verbs from gensim and create hypothesis in the following two ways:
(i) substituting verb with its contradictory verb: for example, from ``\textit{A girl is walking}'', we create hypothesis ``\textit{A girl is driving}'' and
(ii) selecting other sentences from the premise pool that have the same subject as the original sentence but have contradictory verbs: for example, sentences like ``\textit{A young girl is driving fast on the street}'' and ``\textit{There is a girl skiing with her mother}''.
The second approach adds diversity to our synthetically generated PHL triplets\textsuperscript{\ref{footnote1}}. 

\paragraph{Subject Object Swap (SOS):} We swap the subject and object of a sentence to create a contradictory hypothesis. 
For example, from the sentence ``\textit{A clock is standing on top of a concrete pillar}'', we create a contradictory hypothesis ``\textit{a pillar is standing on top of a concrete clock}''.

\paragraph{Negation Introduction (NI):} We introduce negation into a sentence to create a contradictory hypothesis. For example, from the sentence ``\textit{Empty fog covered streets in the night}'', we create hypothesis ``\textit{Empty fog did not cover streets in the night}''.

\paragraph{Number Substitution (NS):} Here, we change numbers (tokens with dependency tag `\textit{nummod}' in the parse tree) in a sentence.
For example, changing `four' to `seven' in the sentence ``\textit{Car has four red lights}'' results in a contradictory hypothesis.

\paragraph{Irrelevant Hypothesis (IrH):} 
We sample sentences that have different subjects and objects than the premise sentence.
For example, for the premise ``\textit{Sign for an ancient monument on the roadside}'', we sample ``\textit{A man goes to strike a tennis ball}'' as a contradictory hypothesis.

\subsubsection{Neutral:}
The label is neutral when the premise does not provide enough information to classify a PH pair as either entailment or contradiction.

\paragraph{Adding Modifiers (AM):} 
We introduce a \textit{relevant} modifier for noun(s) in premise to generate a neutral hypothesis.
For instance, in the sentence ``\textit{A car parked near the fence}'', we insert modifier 'silver' for the noun `car' and create hypothesis ``\textit{A silver car parked near the fence}''.
We collect relevant modifiers for nouns by parsing sentences in the premise pool and  selecting tokens with dependency tag `\textit{amod}' and POS tag `\textit{ADJ}'\textsuperscript{\ref{footnote1}}.





\paragraph{ConceptNet (Con):} We 
add relevant information from ConceptNet~\cite{speer2017conceptnet} relations (`{AtLocation}', `{DefinedAs}', etc.) to the premise and create a neutral hypothesis. 
For instance, from the sentence ``\textit{Bunch of bananas are on a table}'', we create hypothesis``\textit{Bunch of bananas are on a table at kitchen}'' using the `{AtLocation}' relation.

\paragraph{Same Subject but Non-Contradictory Verb (SSNCV)}: 
For a premise, we select sentences from the premise pool that have the same subject as the premise, contain additional noun(s) but no contradictory verbs as neutral hypotheses.
For instance, for premise ``\textit{A small child is sleeping in a bed with a bed cover}'', we sample ``\textit{A child laying in bed sleeping with a chair near by}'' as a hypothesis.

We create more examples by swapping premise and hypothesis of the collected PHL triplets and accordingly change the label.
For instance, swapping $P$ and $H$ in \textbf{HS, ES, etc.} results in neutral examples, swapping $P$ and $H$ in \textbf{AM, Con} results in entailment examples.
Furthermore, we note that transformations \textbf{ES, HS, PS, SOS, NI} result in PH pairs with high word overlap between premise and hypothesis sentences, whereas, transformation \textbf{PA, CV, IrH, SSNCV, etc.} result in PH pairs with low word overlap.
In order to add more diversity to the examples, we use composite transformations on the same sentence such as \textbf{PA} + \textbf{ES} ($L=E$), \textbf{PA} + \textbf{CW} ($L=C$) as shown in Table \ref{tab:transforms_combined}.

\subsection{Data Validation}
\label{human_study}
In order to measure the correctness of our procedurally generated PHL triplets, we validate randomly sampled $50$ instances for each transformation. 
We find that nearly all the instances get correct label assignments in case of \textbf{PA, HS, PS, NI, NS, IrH, AM} transformations. While transformations \textbf{CW, Con, SSNCV} result in a few mislabeled instances. 
Specifically, \textbf{SSNCV} transformation results in the maximum errors ($5$). 
Appendix Section \ref{supp_data_validataion} provides examples of such instances.
While it is beneficial to have noise-free training examples, doing so would require more human effort and increase the data collection cost. Thus, in this work, we study how well we can do solely using the procedurally generated data without investing human effort in either creating instances or eliminating noise.


\section{Training NLI Model}
\label{training_nli_model}
In this section, we describe our approach to develop NLI models for each unsupervised setting.
Table~\ref{tab:none_setting_data_stats} (in Appendix) shows sizes of the generated PHL datasets for each setting.

\subsection{NPH-Setting}
We use the Premise Generation function ($\mathcal{P}$) over raw-text sources, namely, COCO captions, ROC stories, and Wikipedia i.e., $\mathcal{P}(\text{COCO})$, $\mathcal{P}(\text{ROC})$, and $\mathcal{P}(\text{Wiki})$ to compile a set of premises and apply the transformations ($\mathcal{T}$) over them to generate PHL triplets.
We then train a transformer-based 3-class classification model (Section \ref{model}) using the generated PHL triplets for the NLI task.

\subsection{P-Setting}
\label{sec:p_setting}
In this slightly relaxed unsupervised setting, premises of the training dataset are provided.
We directly apply the transformation functions ($\mathcal{T}$) on the given premises and generate PHL triplets. 
Similar to the NPH setting, a 3-class classification model is trained using the generated PHL triplets.

\subsection{PH-Setting}
\label{sec:ph_setting}
In this setting, unlabeled training data is provided.
We present a 2-step approach to develop a model for this setting.
In the first step, we create PHL triplets from the premises and train a model using the generated PHL triplets (same as the P-setting).
In the second step, we \textbf{pseudo-label} the unlabeled PH pairs using the model trained in Step 1.

Here, a naive approach to develop NLI model would be to train using this pseudo-labeled dataset. 
This approach is limited by
confirmation bias i.e overfitting to incorrect pseudo-labels predicted by the model \cite{arazo2020pseudo}.
We address this by filtering instances from the pseudo-labeled dataset based on the model's prediction confidence.
We use the maximum softmax probability (maxProb) as the confidence measure and select only the instances that have high prediction confidence for training the final NLI model.
This approach is based on prior work \cite{hendrycks17baseline} showing that correctly classified examples tend to have greater maximum softmax probabilities than erroneously classified examples.
Furthermore, we investigate two ways of training the final NLI model:
\paragraph{Augmenting with $\mathcal{T}(P)$:}
Train using the selected pseudo-labeled dataset and the PHL triplets generated in Step 1.

\paragraph{Further Fine-tune P-Model:}
Further fine-tune the model obtained in Step 1 with the selected pseudo-labeled dataset instead of fine-tuning one from scratch.


\section{Experiments}

\subsection{Experimental Setup}
\label{exp_details}
\paragraph{Datasets:}
We conduct comprehensive experiments with a diverse set of NLI datasets: SNLI~\cite{bowman-etal-2015-large} (sentence derived from only a single text genre), Multi-NLI~\cite{williams-etal-2018-broad} (sentence derived from multiple text genres), Dialogue NLI~\cite{welleck-etal-2019-dialogue} (sentences from context of dialogues), and Breaking NLI~\cite{glockner-etal-2018-breaking} (adversarial instances).

\paragraph{Model:}
\label{model}
We use BERT-BASE model~\cite{devlin-etal-2019-bert} with a linear layer on top of [CLS] token representation for training the 3-class classification model. 
We trained models for $5$ epochs with a batch sizes of 32 and a learning rate ranging in $\{1{-}5\}e{-}5$. All experiments are done with Nvidia V100 16GB GPUs. 



\paragraph{Baseline Methods:}
We compare our approach with Multimodal Aligned Contrastive Decoupled learning \textbf{(MACD)}~\cite{cui-etal-2020-unsupervised} 
, Single-modal pre-training model \textbf{BERT} \cite{devlin-etal-2019-bert},
Multi-modal pre-training model \textbf{LXMERT} \cite{tan-bansal-2019-lxmert}, and \textbf{VilBert} \cite{NEURIPS2019_c74d97b0}.
 
\subsection{Results}
\label{nli_perf}
\begin{table}[t]
    \centering
    \small 
    \resizebox{\linewidth}{!}{
    \begin{tabular}{p{2cm}p{0.6cm}p{0.6cm}p{0.6cm}p{0.6cm}p{0.6cm}}
    \toprule
        \centering\textbf{Model} & \textbf{SNLI}  & \textbf{MNLI mat.} &  \textbf{MNLI mis.}  & \textbf{DNLI} & \textbf{BNLI}\\
        \midrule
        BERT* & 35.09 & - & - & - & -\\
        LXMERT* & 39.03 & - & - & -& - \\
        VilBert* & 43.13 & - & - & - & -\\
        
        \midrule
        
        $\mathcal{T}(\mathcal{P}(\text{C}))$ & 64.8 &	\textbf{49.01} &	\textbf{50.0}	& \textbf{50.26} & 74.73\\
        
        $\mathcal{T}(\mathcal{P}(\text{R}))$  & 58.51	 & 45.44 &	45.93 & 	47.4 & 67.9  \\
        
        $\mathcal{T}(\mathcal{P}(\text{W}))$ & 55.06 & 	44.15 &	44.25 &	48.48 & 62.58 \\
        $\mathcal{T}(\mathcal{P}(\text{C+R}))$ & \textbf{65.39}	& 46.83	& 46.92	& 47.95 & \textbf{77.37} \\
        $\mathcal{T}(\mathcal{P}(\text{C+R+W}))$ & 65.09 &	46.63 &	46.83 &	44.74 & 56.11 \\

        
    \bottomrule
    \end{tabular}
    }
    \caption{
    Comparing accuracy of models in the \textbf{NPH-setting}. C, R, and W correspond to the premise sources COCO, ROC, and Wikipedia respectively. Results marked with * have been taken from \cite{cui-etal-2020-unsupervised}.
    }
    \label{tab:NPH_perf_table}
\end{table}
\begin{table}[t]
    \centering 
    \small
    \begin{tabular}{@{}p{2cm}p{0.7cm}p{0.7cm}p{0.7cm}p{0.7cm}p{0.7cm}@{}}
    \toprule
        \textbf{Approach} & \textbf{SNLI}  & \textbf{MNLI mat.} &  \textbf{MNLI mis.} & \textbf{DNLI} & \textbf{BNLI} \\
        \midrule
        BERT* & 35.09 & - & - & - & -\\
        LXMERT* & 39.03 & - & - & -& - \\
        VilBert* & 43.13 & - & - & - & -\\
        MACD* & 52.63 & - & - & - & - \\
        \midrule
        
        $\mathcal{T}(\text{SNLI})$ & 65.72 & 49.56 & 50.00 & 43.27 & 67.78\\
        \quad{+}$\mathcal{T}(\mathcal{P}(\text{C}))$ & 65.36 & 49.91 & 49.24 & 46.25 & 70.07\\
        \quad{+}$\mathcal{T}(\mathcal{P}(\text{R}))$ & \textbf{65.90} & 48.53 & 48.36 & 44.97 & 66.43\\
        
    \bottomrule
    \end{tabular}
    \caption{Comparing accuracy of various approaches in the \textbf{P-Setting}.
    Results marked with * have been taken from \cite{cui-etal-2020-unsupervised}.
    Note that we utilize the premises of the SNLI training dataset only but evaluate on SNLI (in-domain), and MNLI, DNLI, BNLI (out-of-domain).
    }
    \label{tab:P_perf_table}
\end{table}
\begin{table}[t]
    \centering
    \resizebox{\linewidth}{!}{
    \begin{tabular}{@{}llp{0.8cm}p{0.8cm}p{0.8cm}}
    \toprule
        \textbf{Method} & \textbf{Data} & \textbf{SNLI}  & \textbf{MNLI mat.} &  \textbf{MNLI mis.} \\
        \midrule
        From Scratch & MaxProbFilt & 66.67 & \textbf{53.37} & \textbf{55.17}\\
        From Scratch & MaxProbFilt{+}$\mathcal{T}(P)$ & \textbf{66.75} & 50.22 & 50.37 \\
        Finetune P-model & MaxProbFilt & 65.60 & 52.97 & 53.44 \\ 
        
    \bottomrule
    \end{tabular}
    }
    \caption{Comparing accuracy of our proposed approaches in the \textbf{PH-Setting}.
    Note that the models are trained using PH pairs only from the SNLI train-set but evaluated on MNLI (out-of-domain dataset) also.}
    \label{tab:PH_perf_table}
\end{table}
\begin{table*}[]
\small
\resizebox{\textwidth}{!}{%
\begin{tabular}{llllllllllll}
\toprule
\multicolumn{1}{c}{\multirow{1}{*}{\textbf{Training}}} &
  \multicolumn{1}{c}{\multirow{2}{*}{\textbf{Method}}}  &
  \multicolumn{2}{c}{\textbf{\underline{100}}} &
  \multicolumn{2}{c}{\textbf{\underline{200}}} &
  \multicolumn{2}{c}{\textbf{\underline{500}}} &
  \multicolumn{2}{c}{\textbf{\underline{1000}}} &
  \multicolumn{2}{c}{\textbf{\underline{2000}}} \\
  
  \multicolumn{1}{c}{\textbf{Dataset}} &
  \multicolumn{1}{l}{} &
  SNLI & MNLI &
  SNLI & MNLI &
  SNLI & MNLI &
  SNLI & MNLI &
  SNLI & MNLI \\
  

\midrule
\multirow{3}{*}{SNLI}
    & BERT & 44.62 & 37.36 &	48.97 & 34.71 &	58.54 & 44.01 &	65.36 & 37.24 &	72.51 & 45.59\\

    & NPH (Random) & \textbf{64.82} & \textbf{49.72} &	\textbf{65.06} & \textbf{50.48} &	\textbf{66.97} & \textbf{52.33} &	\textbf{70.61} & \textbf{56.75} &	\textbf{73.7} & \textbf{59.0} \\
    & NPH (Adv.)  & \textbf{68.21} &	\textbf{51.93} & \textbf{69.23} & \textbf{56.55} &	\textbf{70.85}  & \textbf{58.46} &	\textbf{73.62} & \textbf{59.47} &	\textbf{74.31} & \textbf{60.43} \\

\midrule
\multirow{2}{*}{MNLI}
    & BERT & 35.12 & 36.01 & 35.14	 & 36.58 &	 46.16 &  47.1 & 47.64	 &  56.21 &	53.68 & \textbf{63.3}  \\

    & NPH (Random) & \textbf{63.87} & \textbf{52.85} &	\textbf{63.87} & \textbf{53.61} &	\textbf{64.23} & \textbf{57.47} &	\textbf{65.62} & \textbf{60.42} &	\textbf{66.87} & {62.89} \\


\bottomrule
\end{tabular}
}
\caption{Comparing performance of various methods on in-domain and out-of-domain datasets in \textbf{low-data regimes} (100-2000 training instances). `BERT' method corresponds to fine-tuning BERT over the provided instances from SNLI/MNLI, `NPH (Random)' corresponds to further fine-tuning our NPH model with the randomly sampled instances from SNLI/MNLI, `NPH (Adv.)' corresponds to further fine-tuning our NPH model with the adversarially selected instances from SNLI/MNLI.}
\label{tab:few_shot_results}

\end{table*}



\paragraph{NPH-Setting:} We utilize three raw text sources: COCO, ROC, and Wikipedia to compile a premise pool and then generate PHL triplets from those premises. 
Table \ref{tab:NPH_perf_table} shows the accuracy of models in this setting.
We use equal number of PHL triplets ($150k$ class-balanced) for training the NLI models.
We find that \textbf{the model trained on PHL triplets generated from COCO captions as premises outperforms ROC and Wikipedia models on all datasets}.
We attribute this superior performance to the short, simple, and diverse sentences present in COCO that resemble the premises of SNLI that were collected from Flickr30K \cite{plummer2015flickr30k} dataset.
In contrast, Wikipedia contains lengthy and compositional sentences resulting in premises that differ from those present in SNLI, MNLI, etc.
Furthermore, we find that \textbf{combining the PHL triplets of COCO and ROC leads to a slight improvement in performance on SNLI ($65.39\%$), and BNLI ($77.37\%$) datasets}.

\paragraph{P-Setting:} 
\citet{cui-etal-2020-unsupervised} presented MACD that performs multi-modal pretraining using COCO and Flick30K caption data for the unsupervised NLI task.
It achieves $52.63\%$ on the SNLI dataset.
\textbf{Our approach outperforms MACD and other single-modal and multi-modal baselines by ${\sim}13\%$} on SNLI as shown in Table \ref{tab:P_perf_table}.
We also experiment by adding PHL triplets generated from COCO and ROC to the training dataset that further improves the accuracy to $65.90\%$ and establish a new state-of-the-art performance in this setting.

\paragraph{PH-Setting:} 
Here, we first pseudo-label the given unlabeled PH pairs using the P-model and then select instances based on the maximum softmax probability (Section \ref{sec:ph_setting}). 
We refer to this set of selected instances as \textit{MaxProbFilt} dataset.
This approach results in accuracy of $66.67\%$ on the SNLI dataset as shown in Table \ref{tab:PH_perf_table}.
We investigate two more approaches of training the NLI model.
In the first approach, we train using \textit{MaxProbFilt} and PHL triplets generated from premises.
In the second approach, we further fine-tune the P-model with \textit{MaxProbFilt} dataset.
We find that the first approach slightly improves the accuracy to $66.75\%$.
This also represents our best performance across all the unsupervised settings.
Furthermore, we observe \textbf{improvement in the Out-of-domain datasets also }($53.37\%$ and $55.17\%$ on MNLI matched and mismatched datasets respectively).


\subsection{Low-Data Regimes}
\label{few_shot}

We also conduct experiments in low-data regimes where a few labeled instances are provided.
We select these instances from the training dataset of SNLI/MNLI using the following two strategies:

\paragraph{Random:} 
Here, we randomly select instances from the corresponding training dataset. 
Further fine-tuning our NPH model with the selected instances consistently achieves higher performance than the models fine-tuned from scratch as shown in Table \ref{tab:few_shot_results}.
\textbf{With just $500$ SNLI instances i.e. $\sim0.1\%$ of training dataset, our models achieve $8.4\%$ and $8.32\%$ higher accuracy on SNLI (in-domain) and MNLI (out-of-domain) respectively.}
Furthermore, with $500$ MNLI instances, our models achieve $10.37\%$ and $18.07\%$ higher accuracy on MNLI (in-domain) and SNLI (out-of-domain) respectively.

\paragraph{Adversarial:} 
Here, we select those instances from the training dataset on which the NPH model makes incorrect prediction.
This is similar to the adversarial data collection strategy \cite{nie-etal-2020-adversarial, kiela-etal-2021-dynabench} where instances that fool the model are collected. 
Here, we do not simply fine-tune our NPH model with the adversarial examples as it would lead to catastrophic forgetting \cite{carpenter1988art}.
We tackle this by including $20000$ randomly sampled instances from the generated PHL triplets and fine-tune on the combined dataset.
\textbf{It further takes the performance to $70.85\%$, $58.46\%$ on SNLI and MNLI respectively with $500$ instances.}


\subsection{Analysis}

\paragraph{Ablation Study:}
\begin{table}[t]
    \centering
    \begin{tabular}{@{}lc@{}}
        \toprule
        \textbf{Approach} & \textbf{$\Delta$ Accuracy} \\
        \midrule
        NPH model & $64.8\%$ \\
        \quad{-} CV &  $-5.88\% $  \\
        \quad{-} CW & $-3.07\% $ \\
        \quad{-} SSNCV & $-2.63\% $ \\
        \quad{-} Neg. &  $-0.70\% $ \\
        \quad{-} IrH & $-0.50\% $ \\
        \quad{-} PS & $-0.00\%$ \\
    \bottomrule
    \end{tabular}
    \caption{\textbf{Ablation Study of transformations} in the NPH-Setting. Each row corresponds to the drop in performance on the SNLI dataset when trained without PHL triplets created using that transformation.}
    \label{tab:ablation_study}
\end{table}
\label{ablation_study}
We conduct ablation study to understand the contribution of individual transformations on NLI performance. 
Table~\ref{tab:ablation_study} shows the performance drop observed on removing PHL triplets created using a single transformation in the NPH-Setting.
We find that \textbf{Contradictory Words (CW) and Contradictory Verbs (CV) lead to the maximum drop in performance, $5.88\%$ and $3.07\%$ respectively.}
In contrast, Pronoun Substitution (PS) transformation doesn't impact the performance significantly. 
Note that this does not imply that this transformation is not effective, it means that the evaluation dataset (SNLI) does not contain instances requiring this transformation.

\paragraph{NC and RS Evaluation:}
\label{bias_evaluation}

\begin{table}[t]
    \centering
    \small
    \begin{tabular}{@{}ccccc@{}}
    \toprule
        \textbf{Setting} & \textbf{Metric} & \multicolumn{3}{c}{\textbf{Label}} \\
        \cmidrule{3-5}
        & & \textbf{C}  & \textbf{E} &  \textbf{N}\\
        \midrule
        \multirow{2}{*}{NPH} & Precision  & 0.65 & 0.71 & 0.6  \\
        & Recall & 0.68 & 0.77 & 0.51  \\
        \midrule
        \multirow{2}{*}{P} & Precision & 0.66 & 0.72 & 0.58\\
        & Recall & 0.67 & 0.78 & 0.52\\
        \midrule
        \multirow{2}{*}{PH} & Precision & 0.64 & 0.74 &0.60 \\
        & Recall & 0.73 & 0.77 & 0.50\\
    \bottomrule
    \end{tabular}
    \caption{\textbf{Precision and Recall values} achieved by our models under each unsupervised setting.}
    \label{tab:pr_values}
\end{table}

\begin{table}[t]
    \centering
    \small
    \begin{tabular}{@{}cccc@{}}
    \toprule
        \textbf{NC} & 	\textbf{RS} & 	\textbf{SNLI-RS} & 	\textbf{SNLI-NC} \\
        \midrule
        84.22 &	50.07 &	58.59 &	75.39 \\
        
    \bottomrule
    \end{tabular}
    \caption{Performance of our NPH model on \textbf{Names-Changed (NC) and Roles-Switched (RS) adversarial test sets}~\cite{Mitra2020EnhancingNL}.}
    \label{tab:nothing_table_rs_nc}
\end{table}
We evaluate our model on NER-Changed (NC) and Roles-Switched (RS) datasets presented in~\cite{Mitra2020EnhancingNL} that test the ability to distinguish entities and roles.
\textbf{Our model achieves high performance on these datasets}. 
Specifically, $84.22\%$ on NC and $75.39\%$ on SNLI-NC as shown in Table~\ref{tab:nothing_table_rs_nc}.

\paragraph{Label-Specific Analysis:}
\label{error_analysis}

Table~\ref{tab:pr_values} shows the precision and recall values achieved by our models. 
We observe that our models perform better on Entailment and Contradiction than Neutral examples.
This suggests that \textbf{neutral examples are relatively more difficult.}
We provide examples of instances where our model makes incorrect predictions and conduct error analysis in Appendix.



\section{Conclusion and Discussion}
We explored three different settings in unsupervised NLI and proposed a procedural data generation approach that outperformed the existing unsupervised methods by ${\sim}13$\%.
Then, we showed that fine-tuning our models with a few human-authored instances leads to a considerable improvement in performance.
We also experimented using adversarial instances for this fine-tuning step instead of randomly selected instances and showed that it further improves the performance.
Specifically, in presence of just $500$ adversarial instances, the proposed method achieved $70.85\%$ accuracy on SNLI, $12.2\%$ higher than the model trained from scratch on the same $500$ instances.

This improvement in performance suggests possibility of an alternative data collection strategy that not only results in high-quality data instances but is also resource efficient.
Using a model-in-the-loop technique has been shown to be effective for adversarial data collection \cite{nie-etal-2020-adversarial, kiela-etal-2021-dynabench, li2021adversarial, sheng2021human, arunkumar2020real}.
In these techniques, a model is first trained on a large dataset and then humans are instructed to create adversarial samples that fool the model into making incorrect predictions. 
Thus, requiring the crowd-sourcing effort twice.
However, in our method, a dataset designer can develop a set of simple functions (or transformations) to procedurally generate training data for the model and can directly instruct humans to create adversarial samples to fool the trained model. 
This is resource efficient and allows dataset designers to control the quality of their dataset.


\section*{Ethical Considerations}  
We use existing public-domain text corpora such as Wikipedia, ROC Stories, and MS-COCO, and follow the protocol to use and adapt research data to generate our weakly-labeled dataset. We will release the code to generate our dataset. Any bias observed in NLI systems trained using our methods can be attributed to the source data and our transformation functions. However, no particular socio-political bias is emphasized or reduced specifically by our methods.

\section*{Acknowledgements}
We thank the anonymous reviewers for their insightful feedback. 
This research was supported by DARPA SAIL-ON and DARPA CHESS programs.
The views and opinions of the authors expressed herein do not necessarily state or reflect those of the funding agencies and employers.

\bibliography{anthology,custom}

\begin{thebibliography}{38}
\expandafter\ifx\csname natexlab\endcsname\relax\def\natexlab#1{#1}\fi

\bibitem[{Arazo et~al.(2020)Arazo, Ortego, Albert, O’Connor, and
  McGuinness}]{arazo2020pseudo}
Eric Arazo, Diego Ortego, Paul Albert, Noel~E O’Connor, and Kevin McGuinness.
  2020.
\newblock Pseudo-labeling and confirmation bias in deep semi-supervised
  learning.
\newblock In \emph{2020 International Joint Conference on Neural Networks
  (IJCNN)}, pages 1--8. IEEE.

\bibitem[{Arunkumar et~al.(2020)Arunkumar, Mishra, Sachdeva, Baral, and
  Bryan}]{arunkumar2020real}
Anjana Arunkumar, Swaroop Mishra, Bhavdeep Sachdeva, Chitta Baral, and Chris
  Bryan. 2020.
\newblock Real-time visual feedback for educative benchmark creation: A
  human-and-metric-in-the-loop workflow.

\bibitem[{Banerjee and Baral(2020)}]{banerjee-baral-2020-self}
Pratyay Banerjee and Chitta Baral. 2020.
\newblock \href {https://doi.org/10.18653/v1/2020.emnlp-main.11}
  {Self-supervised knowledge triplet learning for zero-shot question
  answering}.
\newblock In \emph{Proceedings of the 2020 Conference on Empirical Methods in
  Natural Language Processing (EMNLP)}, pages 151--162, Online. Association for
  Computational Linguistics.

\bibitem[{Banerjee et~al.(2021)Banerjee, Gokhale, Yang, and
  Baral}]{banerjee-etal-2021-weaqa}
Pratyay Banerjee, Tejas Gokhale, Yezhou Yang, and Chitta Baral. 2021.
\newblock \href {https://doi.org/10.18653/v1/2021.findings-acl.302} {{W}ea{QA}:
  Weak supervision via captions for visual question answering}.
\newblock In \emph{Findings of the Association for Computational Linguistics:
  ACL-IJCNLP 2021}, pages 3420--3435, Online. Association for Computational
  Linguistics.

\bibitem[{Bowman et~al.(2015)Bowman, Angeli, Potts, and
  Manning}]{bowman-etal-2015-large}
Samuel~R. Bowman, Gabor Angeli, Christopher Potts, and Christopher~D. Manning.
  2015.
\newblock \href {https://doi.org/10.18653/v1/D15-1075} {A large annotated
  corpus for learning natural language inference}.
\newblock In \emph{Proceedings of the 2015 Conference on Empirical Methods in
  Natural Language Processing}, pages 632--642, Lisbon, Portugal. Association
  for Computational Linguistics.

\bibitem[{Brown et~al.(2020)Brown, Mann, Ryder, Subbiah, Kaplan, Dhariwal,
  Neelakantan, Shyam, Sastry, Askell, Agarwal, Herbert-Voss, Krueger, Henighan,
  Child, Ramesh, Ziegler, Wu, Winter, Hesse, Chen, Sigler, Litwin, Gray, Chess,
  Clark, Berner, McCandlish, Radford, Sutskever, and
  Amodei}]{NEURIPS2020_1457c0d6}
Tom Brown, Benjamin Mann, Nick Ryder, Melanie Subbiah, Jared~D Kaplan, Prafulla
  Dhariwal, Arvind Neelakantan, Pranav Shyam, Girish Sastry, Amanda Askell,
  Sandhini Agarwal, Ariel Herbert-Voss, Gretchen Krueger, Tom Henighan, Rewon
  Child, Aditya Ramesh, Daniel Ziegler, Jeffrey Wu, Clemens Winter, Chris
  Hesse, Mark Chen, Eric Sigler, Mateusz Litwin, Scott Gray, Benjamin Chess,
  Jack Clark, Christopher Berner, Sam McCandlish, Alec Radford, Ilya Sutskever,
  and Dario Amodei. 2020.
\newblock \href
  {https://proceedings.neurips.cc/paper/2020/file/1457c0d6bfcb4967418bfb8ac142f64a-Paper.pdf}
  {Language models are few-shot learners}.
\newblock In \emph{Advances in Neural Information Processing Systems},
  volume~33, pages 1877--1901. Curran Associates, Inc.

\bibitem[{Carpenter and Grossberg(1988)}]{carpenter1988art}
Gail~A. Carpenter and Stephen Grossberg. 1988.
\newblock The art of adaptive pattern recognition by a self-organizing neural
  network.
\newblock \emph{Computer}, 21(3):77--88.

\bibitem[{Cui et~al.(2020)Cui, Zheng, and Wang}]{cui-etal-2020-unsupervised}
Wanyun Cui, Guangyu Zheng, and Wei Wang. 2020.
\newblock \href {https://doi.org/10.18653/v1/2020.emnlp-main.444} {Unsupervised
  natural language inference via decoupled multimodal contrastive learning}.
\newblock In \emph{Proceedings of the 2020 Conference on Empirical Methods in
  Natural Language Processing (EMNLP)}, pages 5511--5520, Online. Association
  for Computational Linguistics.

\bibitem[{Devlin et~al.(2019)Devlin, Chang, Lee, and
  Toutanova}]{devlin-etal-2019-bert}
Jacob Devlin, Ming-Wei Chang, Kenton Lee, and Kristina Toutanova. 2019.
\newblock \href {https://doi.org/10.18653/v1/N19-1423} {{BERT}: Pre-training of
  deep bidirectional transformers for language understanding}.
\newblock In \emph{Proceedings of the 2019 Conference of the North {A}merican
  Chapter of the Association for Computational Linguistics: Human Language
  Technologies, Volume 1 (Long and Short Papers)}, pages 4171--4186,
  Minneapolis, Minnesota. Association for Computational Linguistics.

\bibitem[{Dhingra et~al.(2018)Dhingra, Danish, and
  Rajagopal}]{dhingra-etal-2018-simple}
Bhuwan Dhingra, Danish Danish, and Dheeraj Rajagopal. 2018.
\newblock \href {https://doi.org/10.18653/v1/N18-2092} {Simple and effective
  semi-supervised question answering}.
\newblock In \emph{Proceedings of the 2018 Conference of the North {A}merican
  Chapter of the Association for Computational Linguistics: Human Language
  Technologies, Volume 2 (Short Papers)}, pages 582--587, New Orleans,
  Louisiana. Association for Computational Linguistics.

\bibitem[{Fabbri et~al.(2020)Fabbri, Ng, Wang, Nallapati, and
  Xiang}]{fabbri-etal-2020-template}
Alexander Fabbri, Patrick Ng, Zhiguo Wang, Ramesh Nallapati, and Bing Xiang.
  2020.
\newblock \href {https://doi.org/10.18653/v1/2020.acl-main.413} {Template-based
  question generation from retrieved sentences for improved unsupervised
  question answering}.
\newblock In \emph{Proceedings of the 58th Annual Meeting of the Association
  for Computational Linguistics}, pages 4508--4513, Online. Association for
  Computational Linguistics.

\bibitem[{Glockner et~al.(2018)Glockner, Shwartz, and
  Goldberg}]{glockner-etal-2018-breaking}
Max Glockner, Vered Shwartz, and Yoav Goldberg. 2018.
\newblock \href {https://doi.org/10.18653/v1/P18-2103} {Breaking {NLI} systems
  with sentences that require simple lexical inferences}.
\newblock In \emph{Proceedings of the 56th Annual Meeting of the Association
  for Computational Linguistics (Volume 2: Short Papers)}, pages 650--655,
  Melbourne, Australia. Association for Computational Linguistics.

\bibitem[{Gokhale et~al.(2020)Gokhale, Banerjee, Baral, and
  Yang}]{gokhale2020vqa}
Tejas Gokhale, Pratyay Banerjee, Chitta Baral, and Yezhou Yang. 2020.
\newblock Vqa-lol: Visual question answering under the lens of logic.
\newblock In \emph{European conference on computer vision}, pages 379--396.
  Springer.

\bibitem[{Gokhale et~al.(2021)Gokhale, Chaudhary, Banerjee, Baral, and
  Yang}]{gokhale2021semantically}
Tejas Gokhale, Abhishek Chaudhary, Pratyay Banerjee, Chitta Baral, and Yezhou
  Yang. 2021.
\newblock \href {http://arxiv.org/abs/2110.07165} {Semantically distributed
  robust optimization for vision-and-language inference}.

\bibitem[{Hendrycks and Gimpel(2017)}]{hendrycks17baseline}
Dan Hendrycks and Kevin Gimpel. 2017.
\newblock A baseline for detecting misclassified and out-of-distribution
  examples in neural networks.
\newblock \emph{Proceedings of International Conference on Learning
  Representations}.

\bibitem[{Honnibal et~al.(2020)Honnibal, Montani, Van~Landeghem, and
  Boyd}]{spacy}
Matthew Honnibal, Ines Montani, Sofie Van~Landeghem, and Adriane Boyd. 2020.
\newblock \href {https://doi.org/10.5281/zenodo.1212303} {{spaCy:
  Industrial-strength Natural Language Processing in Python}}.

\bibitem[{Kiela et~al.(2021)Kiela, Bartolo, Nie, Kaushik, Geiger, Wu, Vidgen,
  Prasad, Singh, Ringshia, Ma, Thrush, Riedel, Waseem, Stenetorp, Jia, Bansal,
  Potts, and Williams}]{kiela-etal-2021-dynabench}
Douwe Kiela, Max Bartolo, Yixin Nie, Divyansh Kaushik, Atticus Geiger,
  Zhengxuan Wu, Bertie Vidgen, Grusha Prasad, Amanpreet Singh, Pratik Ringshia,
  Zhiyi Ma, Tristan Thrush, Sebastian Riedel, Zeerak Waseem, Pontus Stenetorp,
  Robin Jia, Mohit Bansal, Christopher Potts, and Adina Williams. 2021.
\newblock \href {https://doi.org/10.18653/v1/2021.naacl-main.324} {Dynabench:
  Rethinking benchmarking in {NLP}}.
\newblock In \emph{Proceedings of the 2021 Conference of the North American
  Chapter of the Association for Computational Linguistics: Human Language
  Technologies}, pages 4110--4124, Online. Association for Computational
  Linguistics.

\bibitem[{Lewis et~al.(2019)Lewis, Denoyer, and
  Riedel}]{lewis-etal-2019-unsupervised}
Patrick Lewis, Ludovic Denoyer, and Sebastian Riedel. 2019.
\newblock \href {https://doi.org/10.18653/v1/P19-1484} {Unsupervised question
  answering by cloze translation}.
\newblock In \emph{Proceedings of the 57th Annual Meeting of the Association
  for Computational Linguistics}, pages 4896--4910, Florence, Italy.
  Association for Computational Linguistics.

\bibitem[{Li et~al.(2021)Li, Lei, Gan, and Liu}]{li2021adversarial}
Linjie Li, Jie Lei, Zhe Gan, and Jingjing Liu. 2021.
\newblock Adversarial vqa: A new benchmark for evaluating the robustness of vqa
  models.
\newblock In \emph{International Conference on Computer Vision (ICCV)}.

\bibitem[{Lin et~al.(2014)Lin, Maire, Belongie, Hays, Perona, Ramanan, Dollar,
  and Zitnick}]{lin2014microsoft}
Tsung-Yi Lin, Michael Maire, Serge Belongie, James Hays, Pietro Perona, Deva
  Ramanan, Piotr Dollar, and Larry Zitnick. 2014.
\newblock \href
  {https://www.microsoft.com/en-us/research/publication/microsoft-coco-common-objects-in-context/}
  {Microsoft coco: Common objects in context}.
\newblock In \emph{ECCV}. European Conference on Computer Vision.

\bibitem[{Lu et~al.(2019)Lu, Batra, Parikh, and Lee}]{NEURIPS2019_c74d97b0}
Jiasen Lu, Dhruv Batra, Devi Parikh, and Stefan Lee. 2019.
\newblock \href
  {https://proceedings.neurips.cc/paper/2019/file/c74d97b01eae257e44aa9d5bade97baf-Paper.pdf}
  {Vilbert: Pretraining task-agnostic visiolinguistic representations for
  vision-and-language tasks}.
\newblock In \emph{Advances in Neural Information Processing Systems},
  volume~32. Curran Associates, Inc.

\bibitem[{Miller(1995)}]{miller1995wordnet}
George~A Miller. 1995.
\newblock Wordnet: a lexical database for english.
\newblock \emph{Communications of the ACM}, 38(11):39--41.

\bibitem[{Mitra et~al.(2020)Mitra, Shrivastava, and
  Baral}]{Mitra2020EnhancingNL}
A.~Mitra, Ishan Shrivastava, and Chitta Baral. 2020.
\newblock Enhancing natural language inference using new and expanded training
  data sets and new learning models.
\newblock In \emph{AAAI}.

\bibitem[{Mostafazadeh et~al.(2016)Mostafazadeh, Chambers, He, Parikh, Batra,
  Vanderwende, Kohli, and Allen}]{mostafazadeh-etal-2016-corpus}
Nasrin Mostafazadeh, Nathanael Chambers, Xiaodong He, Devi Parikh, Dhruv Batra,
  Lucy Vanderwende, Pushmeet Kohli, and James Allen. 2016.
\newblock \href {https://doi.org/10.18653/v1/N16-1098} {A corpus and cloze
  evaluation for deeper understanding of commonsense stories}.
\newblock In \emph{Proceedings of the 2016 Conference of the North {A}merican
  Chapter of the Association for Computational Linguistics: Human Language
  Technologies}, pages 839--849, San Diego, California. Association for
  Computational Linguistics.

\bibitem[{Nie et~al.(2020)Nie, Williams, Dinan, Bansal, Weston, and
  Kiela}]{nie-etal-2020-adversarial}
Yixin Nie, Adina Williams, Emily Dinan, Mohit Bansal, Jason Weston, and Douwe
  Kiela. 2020.
\newblock \href {https://doi.org/10.18653/v1/2020.acl-main.441} {Adversarial
  {NLI}: A new benchmark for natural language understanding}.
\newblock In \emph{Proceedings of the 58th Annual Meeting of the Association
  for Computational Linguistics}, pages 4885--4901, Online. Association for
  Computational Linguistics.

\bibitem[{Plummer et~al.(2015)Plummer, Wang, Cervantes, Caicedo, Hockenmaier,
  and Lazebnik}]{plummer2015flickr30k}
Bryan~A Plummer, Liwei Wang, Chris~M Cervantes, Juan~C Caicedo, Julia
  Hockenmaier, and Svetlana Lazebnik. 2015.
\newblock Flickr30k entities: Collecting region-to-phrase correspondences for
  richer image-to-sentence models.
\newblock In \emph{Proceedings of the IEEE international conference on computer
  vision}, pages 2641--2649.

\bibitem[{Puri et~al.(2020)Puri, Spring, Shoeybi, Patwary, and
  Catanzaro}]{puri-etal-2020-training}
Raul Puri, Ryan Spring, Mohammad Shoeybi, Mostofa Patwary, and Bryan Catanzaro.
  2020.
\newblock \href {https://doi.org/10.18653/v1/2020.emnlp-main.468} {Training
  question answering models from synthetic data}.
\newblock In \emph{Proceedings of the 2020 Conference on Empirical Methods in
  Natural Language Processing (EMNLP)}, pages 5811--5826, Online. Association
  for Computational Linguistics.

\bibitem[{Radford et~al.(2019)Radford, Wu, Child, Luan, Amodei, and
  Sutskever}]{radford2019language}
Alec Radford, Jeffrey Wu, Rewon Child, David Luan, Dario Amodei, and Ilya
  Sutskever. 2019.
\newblock Language models are unsupervised multitask learners.
\newblock \emph{OpenAI blog}, 1(8):9.

\bibitem[{Rehurek and Sojka(2011)}]{rehurek2011gensim}
Radim Rehurek and Petr Sojka. 2011.
\newblock Gensim--python framework for vector space modelling.
\newblock \emph{NLP Centre, Faculty of Informatics, Masaryk University, Brno,
  Czech Republic}, 3(2).

\bibitem[{Sennrich et~al.(2016)Sennrich, Haddow, and
  Birch}]{sennrich-etal-2016-improving}
Rico Sennrich, Barry Haddow, and Alexandra Birch. 2016.
\newblock \href {https://doi.org/10.18653/v1/P16-1009} {Improving neural
  machine translation models with monolingual data}.
\newblock In \emph{Proceedings of the 54th Annual Meeting of the Association
  for Computational Linguistics (Volume 1: Long Papers)}, pages 86--96, Berlin,
  Germany. Association for Computational Linguistics.

\bibitem[{Sheng et~al.(2021)Sheng, Singh, Goswami, Magana, Galuba, Parikh, and
  Kiela}]{sheng2021human}
Sasha Sheng, Amanpreet Singh, Vedanuj Goswami, Jose Alberto~Lopez Magana,
  Wojciech Galuba, Devi Parikh, and Douwe Kiela. 2021.
\newblock Human-adversarial visual question answering.

\bibitem[{Speer et~al.(2017)Speer, Chin, and Havasi}]{speer2017conceptnet}
Robyn Speer, Joshua Chin, and Catherine Havasi. 2017.
\newblock Conceptnet 5.5: An open multilingual graph of general knowledge.
\newblock In \emph{Proceedings of the AAAI Conference on Artificial
  Intelligence}, volume~31.

\bibitem[{Tan and Bansal(2019)}]{tan-bansal-2019-lxmert}
Hao Tan and Mohit Bansal. 2019.
\newblock \href {https://doi.org/10.18653/v1/D19-1514} {{LXMERT}: Learning
  cross-modality encoder representations from transformers}.
\newblock In \emph{Proceedings of the 2019 Conference on Empirical Methods in
  Natural Language Processing and the 9th International Joint Conference on
  Natural Language Processing (EMNLP-IJCNLP)}, pages 5100--5111, Hong Kong,
  China. Association for Computational Linguistics.

\bibitem[{Wang et~al.(2019)Wang, Pruksachatkun, Nangia, Singh, Michael, Hill,
  Levy, and Bowman}]{Wang2019SuperGLUEAS}
Alex Wang, Yada Pruksachatkun, Nikita Nangia, Amanpreet Singh, Julian Michael,
  Felix Hill, Omer Levy, and Samuel~R. Bowman. 2019.
\newblock Superglue: A stickier benchmark for general-purpose language
  understanding systems.
\newblock In \emph{NeurIPS}.

\bibitem[{Wang et~al.(2021)Wang, Yu, Firat, and Cao}]{wang2021towards}
Zirui Wang, Adams~Wei Yu, Orhan Firat, and Yuan Cao. 2021.
\newblock Towards zero-label language learning.
\newblock \emph{arXiv preprint arXiv:2109.09193}.

\bibitem[{Welleck et~al.(2019)Welleck, Weston, Szlam, and
  Cho}]{welleck-etal-2019-dialogue}
Sean Welleck, Jason Weston, Arthur Szlam, and Kyunghyun Cho. 2019.
\newblock \href {https://doi.org/10.18653/v1/P19-1363} {Dialogue natural
  language inference}.
\newblock In \emph{Proceedings of the 57th Annual Meeting of the Association
  for Computational Linguistics}, pages 3731--3741, Florence, Italy.
  Association for Computational Linguistics.

\bibitem[{Williams et~al.(2018)Williams, Nangia, and
  Bowman}]{williams-etal-2018-broad}
Adina Williams, Nikita Nangia, and Samuel Bowman. 2018.
\newblock \href {https://doi.org/10.18653/v1/N18-1101} {A broad-coverage
  challenge corpus for sentence understanding through inference}.
\newblock In \emph{Proceedings of the 2018 Conference of the North {A}merican
  Chapter of the Association for Computational Linguistics: Human Language
  Technologies, Volume 1 (Long Papers)}, pages 1112--1122, New Orleans,
  Louisiana. Association for Computational Linguistics.

\bibitem[{Zhang et~al.(2019)Zhang, Zhao, Saleh, and Liu}]{zhang2019pegasus}
Jingqing Zhang, Yao Zhao, Mohammad Saleh, and Peter~J. Liu. 2019.
\newblock \href {http://arxiv.org/abs/1912.08777} {Pegasus: Pre-training with
  extracted gap-sentences for abstractive summarization}.

\end{thebibliography}
\bibliographystyle{acl_natbib}

\appendix
\section*{Appendix}
\section{Transformations}
\label{supp_transformations}
In this section, we provide details about the proposed sentence transformations.

\subsection{Entailment}

Table \ref{tab:transforms_entailment} shows examples of our transformations.
\paragraph{Paraphrasing (PA):} 
It is an effective way of creating entailment examples as the hypothesis which is simply a paraphrased version of the premise is always entailed
Furthermore, since the Pegasus tool is trained for abstractive text summarization, it often removes some information from the original sentence while paraphrasing. For instance,  a paraphrase of the sentence ``\textit{A boy is playing with a red ball}" could be ``\textit{Boy is playing with a ball}". This restricts us from using the paraphrased sentence as the premise with the original sentence as the hypothesis as the formed $PH$ pair does not represent an entailment scenario (neutral in this case). 
It is non-trivial to detect such instances in an automated way. Hence, in order to avoid noisy examples, we only use the original sentence as premise and paraphrased sentences as hypothesis.
We also explore back-translation \cite{sennrich-etal-2016-improving} but it often results in noisy outputs and provides less diversity than the Pegasus tool. 
Hence, we use only the Pegasus tool for generating paraphrases of sentences.

\paragraph{Extracting Snippets (ES):}
Here, we provide details of the techniques used for extracting snippets from a text. 
Note that we use dependency parse tree of the sentence to select/skip the tokens to create the hypothesis.

(i) We skip modifiers (tokens with dependency \textbf{amod}) that have no children in the parse tree. For example, from the sentence ``\textit{The male surfer is riding a small wave}'', we create ``\textit{The surfer is riding a small wave}'', ``\textit{The male surfer is riding a wave}'', and ``\textit{The surfer is riding a wave}'' as entailing hypotheses.

(ii) Similar to the previous technique, we skip adverb modifier (\textbf{advmod}). For example, from the sentence ``\textit{A very beautiful girl is standing outside the park}'', we create an entailment hypothesis ``\textit{A beautiful girl is standing outside the park}''.

(iii) We skip adjectives that do not have dependency token \textbf{conj} and also have 0 children in the parse tree. For example, from the sentence ``\textit{A middle-aged man in a beige vest is sleeping on a wooden bench.}'', we create ``\textit{A middle-aged man in a vest is sleeping on a bench.}''.

(iv) In another technique, we select the root token and all the tokens to the left of it. If this results in selection of at least 3 tokens and if one of them is a verb then we consider it to be a valid sentence and use it as an entailing hypothesis. For example, from the sentence ``\textit{The male surfer is riding a small wave}'', we create ``\textit{surfer is riding}''.

\paragraph{Hypernym Substitution (HS):}
Examples of hypernyms:

`alcohol': [`beverage', `drink']

`apple': [`fruit']

`axe': [`edge tool']

`banana': [`fruit']

etc.

\paragraph{Pronoun Substitution (PS):}
For words in the list [`man', `boy', `guy', `lord', `husband', `father', `boyfriend', `son', `brother', `grandfather', `uncle'], we use (`he'/ `someone'/ `they', etc.) and for words in the list [`woman', `girl', `lady', `wife', `mother', `daughter', `sister', `girlfriend', `grandmother', `aunt'], we use `she'/ `someone'/ `they', etc.). In other cases, we use the pronoun `they' or `someone' or `somebody'.

\paragraph{Counting (CT):}
We provide examples of templates we use to create counting hypotheses:

``\textit{There are \{count\} \{hypernym\} present}'',

``\textit{\{count\} \{hypernym\} are present}'',

``\textit{Several \{hypernym\} present}'',

``\textit{There are multiple \{hypernym\} present}'',

``\textit{There are more than \{count'\} \{hypernym\} present}'',

``\textit{There are at least \{count'\} \{hypernym\} present}'',

etc.

We also substitute the hypernym in the original sentence directly to create hypotheses as shown in Table \ref{tab:transforms_entailment}.

\subsection{Contradiction}

Table \ref{tab:transforms_contradiction} shows examples of our transformations.
\paragraph{Contradictory Words (CW):}
For contradictory adjectives, we collect antonyms from wordnet and for contradictory nouns, we use the function `$most\_similar$' from gensim \cite{rehurek2011gensim} library.
that returns words close (but distinct) to a given word\textsuperscript{\ref{footnote1}}.
For instance, it returns words like 'piano', 'flute', 'saxophone' when given the word 'violin'
In order to filter out the inflected forms of the same word or its synonyms from the list returned by $most\_similar$ function, we remove words that have high STS with the given word. This step removes noisy contradictory word pairs to a large extent.
Here, we provide examples of contradictory words:

`stove': [`heater']

`cucumber':	[`onion', `carrot', `melon', `turnip',  `eggplant', `watermelon', `radish']

`motorcycle': [`truck', `scooter', `car']

`kitchen':	[`bedroom', `bathroom', `toilet']

etc.

\paragraph{Contradictory Verb (CV):} 
We provide examples of contradictory verbs:

`stand': [`sprint', `cycle', `drive', `jump', `sit', etc.]

`play':[`sleep', `cry', `fight', `drink', `hunt', etc.]

`smile': [`cry', `anger', `frown', etc.]

etc.

\subsection{Neutral}

Table \ref{tab:transforms_neutral} shows examples of our transformations.
\paragraph{Adding Modifiers (AM):}
We provide examples of modifiers collected using our approach:

`metal': [`large', `circular', `galvanized',`heavy', `dark', etc.]

`vegetable': [`steamed', `cruciferous', `green', `uncooked', `raw', etc.]

`park': [`quiet', `neglected', `vast', `square', `crowded', etc.]

etc.
    
\paragraph{ConceptNet:} We use ConceptNet relations \textit{AtLocation}, \textit{DefinedAs}, etc. and insert the node connected by these relations to the sentence resulting in a neutral hypothesis.

\clearpage 
\begin{table*}[t]
\small
    \centering
    \resizebox{\linewidth}{!}{
    \begin{tabular}{@{}p{0.1\linewidth}>{\RaggedRight}p{0.4\linewidth}>{\RaggedRight}p{0.4\linewidth}@{}}
    \toprule
        \textbf{Category} &
        \textbf{Original Sentence (Premise)} &
        \textbf{Hypothesis} 
        \\
    \midrule
          PA	& 
          Fruit and cheese sitting on a black plate.	& 
          There is fruit and cheese on a black plate.	\\
          ES &
          person relaxes at home while holding something.	&
          person relaxes while holding something.	\\
          HS. &	
          A girl is sitting next to a blood hound.	&
          A girl is sitting next to an animal.	\\
          PS	&
          People are walking down a busy city street.	&
          they are walking down a busy city street 	\\
        CT & 
        A man and woman setup a camera. & 
        Two people setup a camera \\
        
        Composite & 
        A large elephant is very close to the camera. &
        elephant is close to the photographic equipment.\\
       
    \bottomrule

    \end{tabular}
    }
    \caption{
    Illustrative examples of entailment transformations. 
    }
    \label{tab:transforms_entailment}
\end{table*}
\begin{table*}[t]
\small
    \centering
    \resizebox{\linewidth}{!}{
    \begin{tabular}{@{}p{0.1\linewidth}>{\RaggedRight}p{0.5\linewidth}>{\RaggedRight}p{0.4\linewidth}@{}}
    \toprule
        \textbf{Category} &
        \textbf{Original Sentence (Premise)} &
        \textbf{Hypothesis}
        \\
    \midrule
         
        CW-noun & 
        A small bathroom with a sink under a cabinet. & 
        a small kitchen with a  sink under a cabinet.\\ 
        
        CW-adj & 
        A young man is doing a trick on a surfboard. & 
        A old man is doing a trick on a surfboard. \\ 
        
        CV & 
        A couple pose for a picture while standing next to a couch. & 
        A couple sit in a chair on laptops \\ 
        
        SOS & 
        A man is flying a kite on the beach. & 
        a beach is flying a kite on the man \\ 
        
        NS & 
        Two green traffics lights in a European city. & 
        nine green traffics lights in a European city\\ 
        
        IrH. & 
        A flock of sheep grazing in a field. & 
        A man having fun as he glides across the water. \\ 
        
        NI. & 
        A boy with gloves on a field throwing a ball. & 
        a boy with gloves on a field not throwing a ball \\ 
        
        Composite &
        A woman holding a baby while a man takes a picture of them &
        a kid is taking a picture of a male and a baby.\\
    \bottomrule

    \end{tabular}
    }
    \caption{Illustrative examples of contradiction transformations.}
    \label{tab:transforms_contradiction}
\end{table*}
\begin{table*}[t]
    \small
    \centering
    \resizebox{\linewidth}{!}{
    \begin{tabular}{@{}p{0.1\linewidth}>{\RaggedRight}p{0.5\linewidth}>{\RaggedRight}p{0.4\linewidth}@{}}
    \toprule
        \textbf{Category} &
        \textbf{Original Sentence (Premise)} &
        \textbf{Hypothesis} 
        \\
    \midrule
         
        AM & 
        two cats are eating next to each other out of the bowl  & 
        two cats are eating next to each other out of the same bowl \\ 
        
        
        SSNCV & 
        A man holds an electronic device over his head. & 
        man is taking photo with a small device \\ 
        
        FCon & 
        a food plate on a table with a glass. & 
        a food plate on a table with a glass which is made of plastic.\\ 
        
        Composite &
        two dogs running through the snow. &
        The big dogs are outside. \\
    \bottomrule

    \end{tabular}
    }
    \caption{Illustrative examples of neutral transformations. }
    \label{tab:transforms_neutral}
\end{table*}

\begin{table*}[t]
    \small
    \centering
    \resizebox{\linewidth}{!}{
    \begin{tabular}{@{}p{0.06\linewidth}>{\RaggedRight}p{0.35\linewidth}>{\RaggedRight}p{0.35\linewidth}>{\RaggedRight}p{0.08\linewidth}>{\RaggedRight}p{0.08\linewidth}@{}}
    \toprule
        \textbf{Trans.} &
        \centering\textbf{Premise} &
        \textbf{Hypothesis} &
        \textbf{Assigned Label} &
        \textbf{True Label}
        \\
    \midrule
        
        PS	& 
        Two dogs on leashes sniffing each other as people walk in a outdoor market &
        Two dogs on leashes sniffing each other as they walk in a market &
        E  &
        N \\
        
        CT	& 
        Adult woman eating slice of pizza while standing next to building &
        There are 2 humans present &
        E  &
        C \\
        
        CW	& 
        Meal with meat and vegetables served on table &
        There is a meal with cheese and vegetables &
        C  &
        N \\
        
        SSNCV &
        A person riding skis down a snowy slope &
        A person riding skis in a body of water &
        N &
        C \\
        
        SSNCV &
        A person on a skateboard jumping up into the air &
        A person jumping up in the air on a snowboard &
        N &
        C \\
        
        CV &
        A male surfer riding a wave on the ocean &
        A surfer is surfing in the ocean near some swimmers &
        C &
        N \\
        
    \bottomrule
    \end{tabular}
    }
    \caption{Examples of mis-labeled PHL triplets generated by our transformations.}
    \label{tab:mislabeled_examples}
\end{table*}

\clearpage
\section{Data Validation}
\label{supp_data_validataion}
Table \ref{tab:mislabeled_examples} shows examples of mis-labeled instances generated by our transformations.

\section{Training NLI Model}
\begin{table}[t]
    \small
    \centering
    \resizebox{\linewidth}{!}{
    \begin{tabular}{@{}>{\RaggedRight}p{2.2cm}>{\RaggedRight}p{0.8cm}>{\RaggedRight}p{0.8cm}>{\RaggedRight}p{0.8cm}>{\RaggedRight}cc@{}}
    \toprule
        \multirow{2}{*}{Transformation $\mathcal{T}$} & \multicolumn{3}{c}{NPH-Setting} & \hphantom &  P-Setting\\
        \cmidrule{2-4} \cmidrule{6-6}
         & $\mathcal{T}(\mathcal{P}(\text{C}))$  & $\mathcal{T}(\mathcal{P}(\text{R}))$ &  $\mathcal{T}(\mathcal{P}(\text{W}))$ && $\mathcal{T}(\text{SNLI})$\\
        \toprule
        Raw Sentences & 591 & 490 & 600 && 548 \\
        \midrule
        PA &  5083 & 3072 & 273 && 475 \\
        ES & 2365 & 196 & 87 && 516\\
        PS & 37 & 41 & 137 && 38\\
        CT & 25 & 8 & 2 && 43\\
        Neg. & 1175 & 1175 & 2053 && 990\\
        CW & 978 & 119 & 116 && 265\\
        CV & 1149 & 63 & 5 && 505\\
        NS & 73 & 16 & 224 && 91\\
        SOS  & 428 & 180 & 229 && 76\\
        AM & 1048 & 125 & 535 && 327\\
        SSNCV & 1363 & 2 & 7 && 405\\
    \bottomrule
    \end{tabular}
    }
    \caption{
    Sizes of PHL triplet datasets generated by our transformations for the unsupervised settings.
    All numbers are in thousands. 
    C, R, W denote COCO, ROC Stories, and Wikipedia respectively. For P-Setting, we show stats for SNLI dataset. We do not include PH-Setting in this table because we leverage the PHL triplets generated using the P-Setting to solve it as described in Section \ref{sec:ph_setting}.}
    \label{tab:none_setting_data_stats}
\end{table}

Table~\ref{tab:none_setting_data_stats} shows sizes of the generated PHL datasets for each setting.




\end{document}